\documentclass[preprint]{elsarticle}

\usepackage{graphicx}

\usepackage{amssymb}

\usepackage{lineno}

\usepackage{comment}
\usepackage{amsmath}
\usepackage{lscape}
\usepackage{multicol}
\usepackage{array}
\usepackage{txfonts}
\usepackage{multirow}
\usepackage{url}
\usepackage{float}
\newcolumntype{P}[1]{>{\centering\arraybackslash}p{#1}}
\newcolumntype{M}[1]{>{\centering\arraybackslash}m{#1}}

\usepackage[pdftex]{color}
\usepackage{xcolor}
\usepackage{colortbl}
\definecolor{Gray}{gray}{0.7}
\definecolor{Blue}{HTML}{0000FF}
\definecolor{Orange}{HTML}{FFD4A1}
\usepackage[font=footnotesize,labelfont=bf]{caption}
\usepackage[font=footnotesize,labelfont=bf]{subcaption}
\usepackage{hyperref}

\journal{}

\begin{document}

\begin{frontmatter}


\title{Adaptive Target Localization under Uncertainty using Multi-Agent Deep Reinforcement Learning with Knowledge Transfer}

\author[add1]{Ahmed Alagha}
\author[add2,add3]{Rabeb Mizouni}
\author [add2,add3]{Shakti Singh}
\author[add2,add4,add1]{Jamal Bentahar}
\author [add2,add3]{Hadi Otrok}

\address[add1]{Concordia Institute for Information Systems Engineering, Concordia University, Montreal, QC, Canada}
\address[add2]{Department of Computer Science, Khalifa University, Abu Dhabi, UAE}
\address [add3]{Center of Cyber Physical Systems (C2PS), Khalifa University, Abu Dhabi, UAE}
\address [add4]{6G Research Center, Khalifa University, Abu Dhabi, UAE}
 
\begin{abstract}
Target localization is a critical task in sensitive applications, where multiple sensing agents communicate and collaborate to identify the target location based on sensor readings. Existing approaches investigated the use of Multi-Agent Deep Reinforcement Learning (MADRL) to tackle target localization. Nevertheless, these methods do not consider practical uncertainties, like false alarms when the target does not exist or when it is unreachable due to environmental complexities. To address these drawbacks, this work proposes a novel MADRL-based method for target localization in uncertain environments. The proposed MADRL method employs Proximal Policy Optimization to optimize the decision-making of sensing agents, which is represented in the form of an actor-critic structure using Convolutional Neural Networks. The observations of the agents are designed in an optimized manner to capture essential information in the environment, and a team-based reward functions is proposed to produce cooperative agents. The MADRL method covers three action dimensionalities that control the agents' mobility to search the area for the target, detect its existence, and determine its reachability. Using the concept of Transfer Learning, a Deep Learning model builds on the knowledge from the MADRL model to accurately estimating the target location if it is unreachable, resulting in shared representations between the models for faster learning and lower computational complexity. Collectively, the final combined model is capable of searching for the target, determining its existence and reachability, and estimating its location accurately. The proposed method is tested using a radioactive target localization environment and benchmarked against existing methods, showing its efficacy.
\end{abstract}

\end{frontmatter}


\section{Introduction}
\label{Sec: Intro}
Target localization is a task concerned with determining the precise location of a specific target within a certain environment. It is a fundamental task in many applications related to rescue missions \cite{de2024safe}, object tracking \cite{hussain2023predictive}, radiation monitoring \cite{alagha2019data, alagha2022target}, and fire detection \cite{ramadan2024towards}. The recent advancements in robot and UAV swarms and their sensing and communication capabilities have revolutionized target localization tasks, enabling more efficient, accurate, and scalable solutions \cite{salameh2023federated, charef2023artificial, grassi2023emergency}. In such systems, the sensing agents cooperate to 
find the unknown location of a target using their observations. The complexity of target localization tasks is represented by the challenges inherent in its sensitive applications, where obstacles can impede the movement and attenuate sensor readings of the agents, the targets may be unreachable, and false warnings can occur \cite{alagha2023multi, liu2019double}.    

While initial target localization methods relied on data fusion techniques to estimate the target location \cite{bai2014maximum, chin2008accurate}, recent methods use Deep Reinforcement Learning (DRL) to develop intelligent agents capable of moving and searching for the target \cite{liu2019double, proctor2021proximal, alagha2022target, alagha2023multi, shurrab2023reinforcement}. DRL offers better adaptability to varied complex environments as well as intelligent decision-making that relies on more than mere data readings. In DRL, a sensing agent builds its own intelligence through rewarded interactions with the environment \cite{sutton2018reinforcement}. Using a pre-designed DRL method, the agent trains a decision-making policy with the collected experiences and rewards, which can then be deployed in a real-life scenario. Multi-agent Deep Reinforcement Learning (MADRL) extends DRL by allowing multiple agents to coexist and learn to cooperate or compete within the same environment. In the realm of target localization, the works in \cite{liu2019double, proctor2021proximal} proposed DRL solutions in single-agent settings, which are inefficient when extended to multi-agent settings due to the curse of dimensionality that results in scalability issues \cite{nguyen2020deep}. Other works \cite{alagha2022target, alagha2023multi} offer MADRL-based solutions for target localization assuming the target always exists and is reachable. However, in realistic scenarios, uncertainties about the existence and reachability of the target arise. In continuous surveillance systems, sensing agents must be able to determine if the target exists before wasting resources on the search. Additionally, in many cases, the target might be unreachable due to environment complexities such as obstacles, necessitating agents to detect such a scenario and provide an alternative target location estimate. The above existing proposals fail to consider the existence or reachability of the target, rendering them infeasible for realistic scenarios.

To overcome the above challenges, this work introduces a novel MADRL-based approach for target localization in uncertain environments. This approach accounts for the target's reachability and existence as well as the potential need to estimate the target location if unreachable. The primary objective is to develop an intelligent team of cooperative sensing agents that can translate their observations into actions that efficiently and quickly find the target location with minimal resource exploitation. The agents' decision-making process involves three key dimensions: mobility, target existence detection, and reachability determination. This is accomplished using a MADRL method based on Proximal Policy Optimization (PPO) \cite{schulman2017proximal}, which trains Convolutional Neural Networks (CNNs) to serve as the agents' decision-making policies. Observations by the agents are represented as 2D heatmaps to capture spatio-temporal dependencies in the collected data. A shaped team-based reward function is proposed, which considers the different uncertainties and ensures fast learning convergence, full cooperation between the agents, and fast localization with low resource consumption. To ensure scalable learning, a centralized learning and distributed execution (CLDE) method is adopted, which mitigates the curse of dimensionality issue in MADRL. If the target is determined to be unreachable, the agents estimate its location using a Deep Learning (DL) model enhanced by Transfer Learning (TL). The DL model extrapolates the feature extractor of the MADRL model using TL, producing a final combined model with a shared feature extractor and two output heads; one for continuous decision-making and one for target estimation, thus reducing computational overhead. The key contributions of this work are summarized as:

\begin{enumerate}
    \item The formulation of the target localization problem in uncertain environments using MADRL.
    \item The design of the agents' observations in an optimized manner that captures the essential information from the environment for optimal training. 
    \item The design of a reward function and decision-making actions over three dimensionalities that determine agents' mobility, target detection, and target reachability.
    \item The development of a learning framework based on TL that leverages the knowledge of the trained MADRL model in training a DL model for target estimation, allowing for reduced computational overhead.
\end{enumerate}

The proposed method is assessed within the context of radiation localization, in which a team of sensing agents is tasked with finding the location of a radioactive material by intelligently searching the area. The proposed work is compared with multiple target localization benchmarks \cite{ziock2002lost, liu2010analysis, xiao2017sampling, sartoretti2019primal, alagha2022target, alagha2023multi} in terms of the learning performance and target localization efficiency.

\section{Problem Definition}
\label{sec: Problem Formulation}

In target localization problems, a team of $N$ agents is deployed in an Area of Interest (AoI) with the task of cooperating to find the unknown target location in a timely and resource-efficient manner. Each agent is equipped with sensors that help acquire information (observations) about the target, the environment, and the other agents. The observations are acquired progressively and utilized for decision-making to take actions in the environment, with the aim of reaching the unknown target location. At each time step, the agents update their observations and use the updated information to act in the environment. It is assumed that agents can communicate and share information, as well as store previously collected observations.

\begin{figure}[h]
     \centering
     \begin{subfigure}{0.32\columnwidth}
         \centering
         \includegraphics[width=\linewidth]{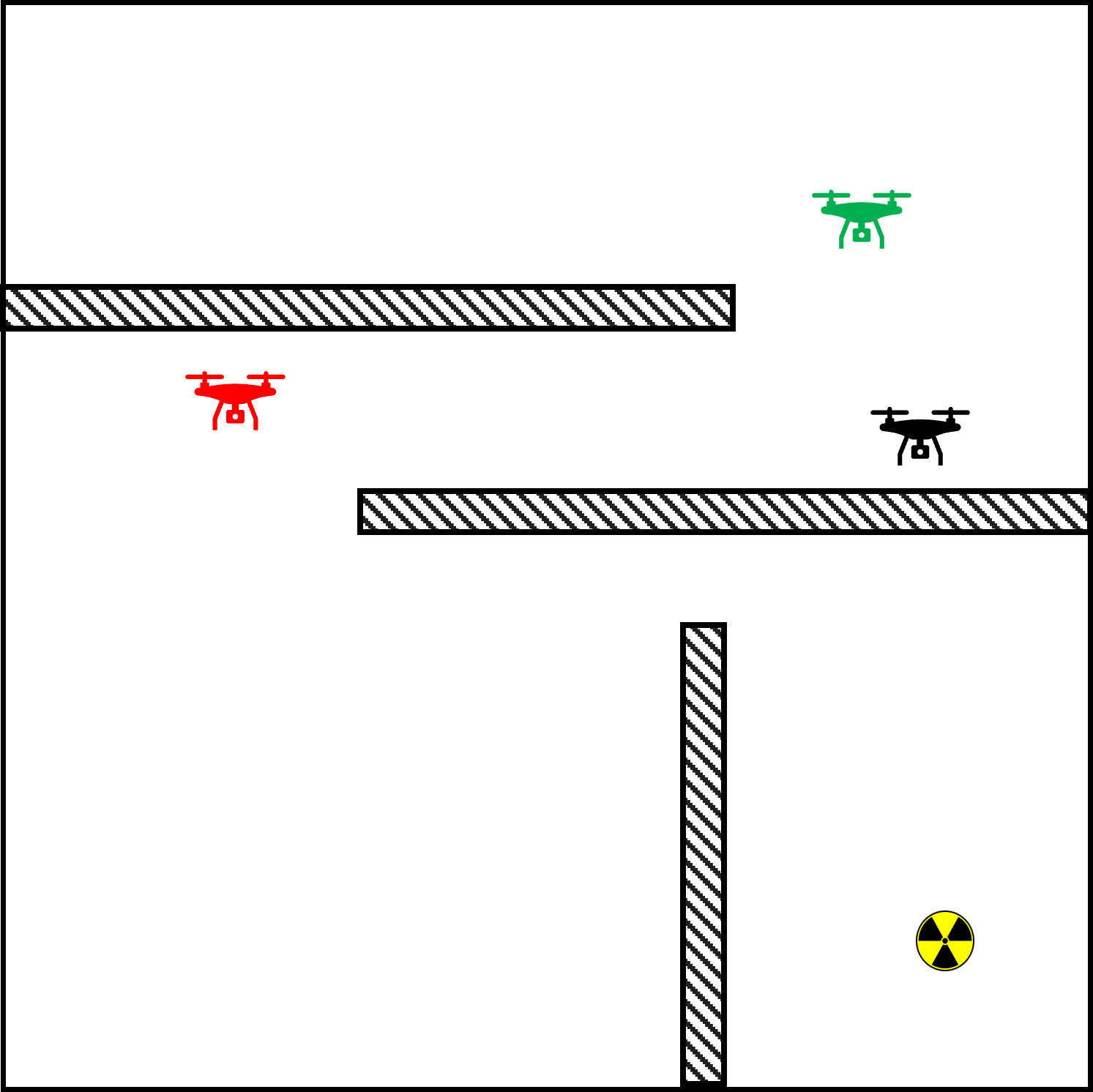}
         \caption{}
         \label{FigScenario1}
     \end{subfigure}
     \begin{subfigure}{0.32\columnwidth}
         \centering
         \includegraphics[width=\linewidth]{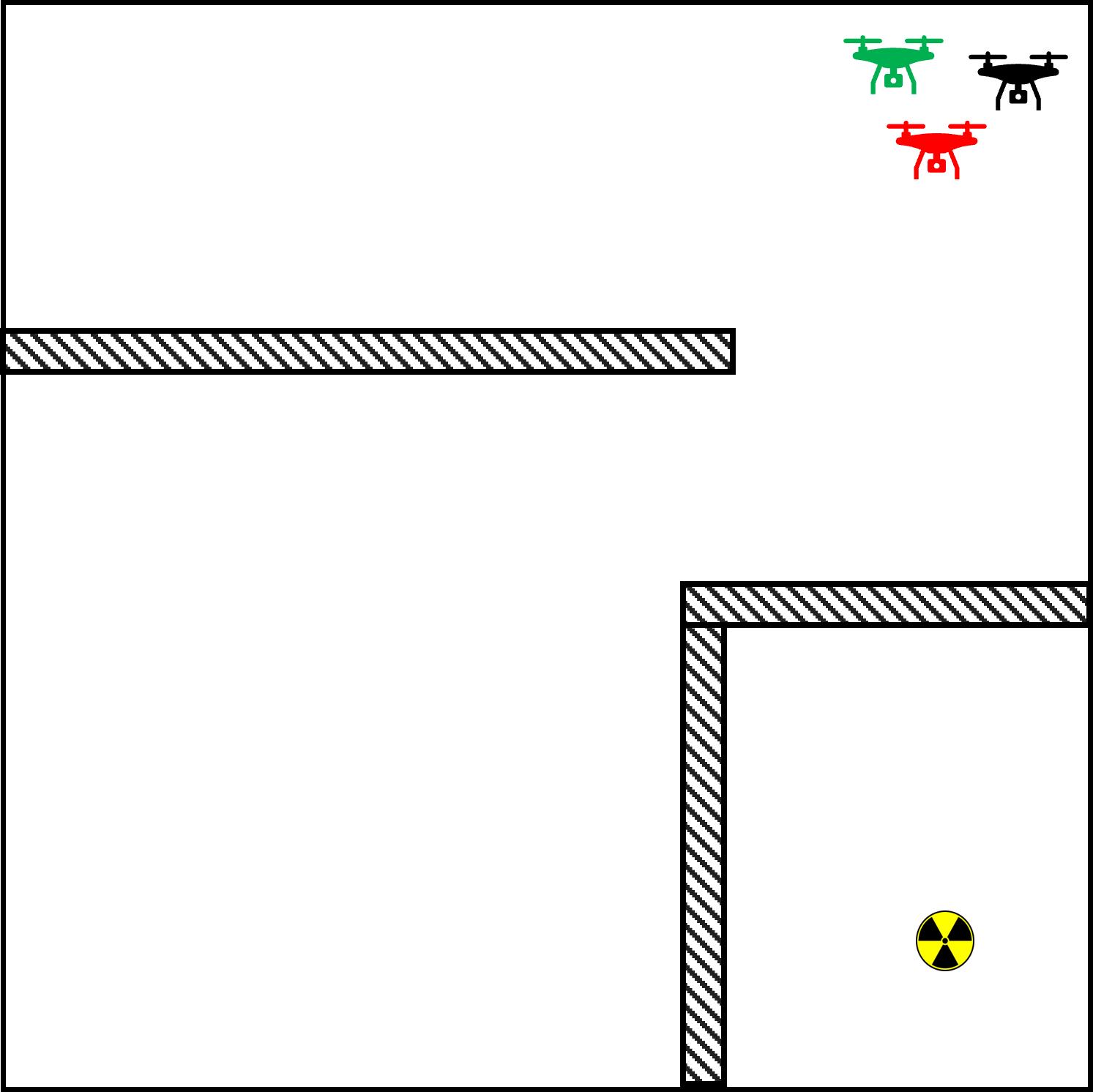}
         \caption{}
         \label{FigScenario2}
     \end{subfigure}
     \begin{subfigure}{0.32\columnwidth}
         \centering
         \includegraphics[width=\linewidth]{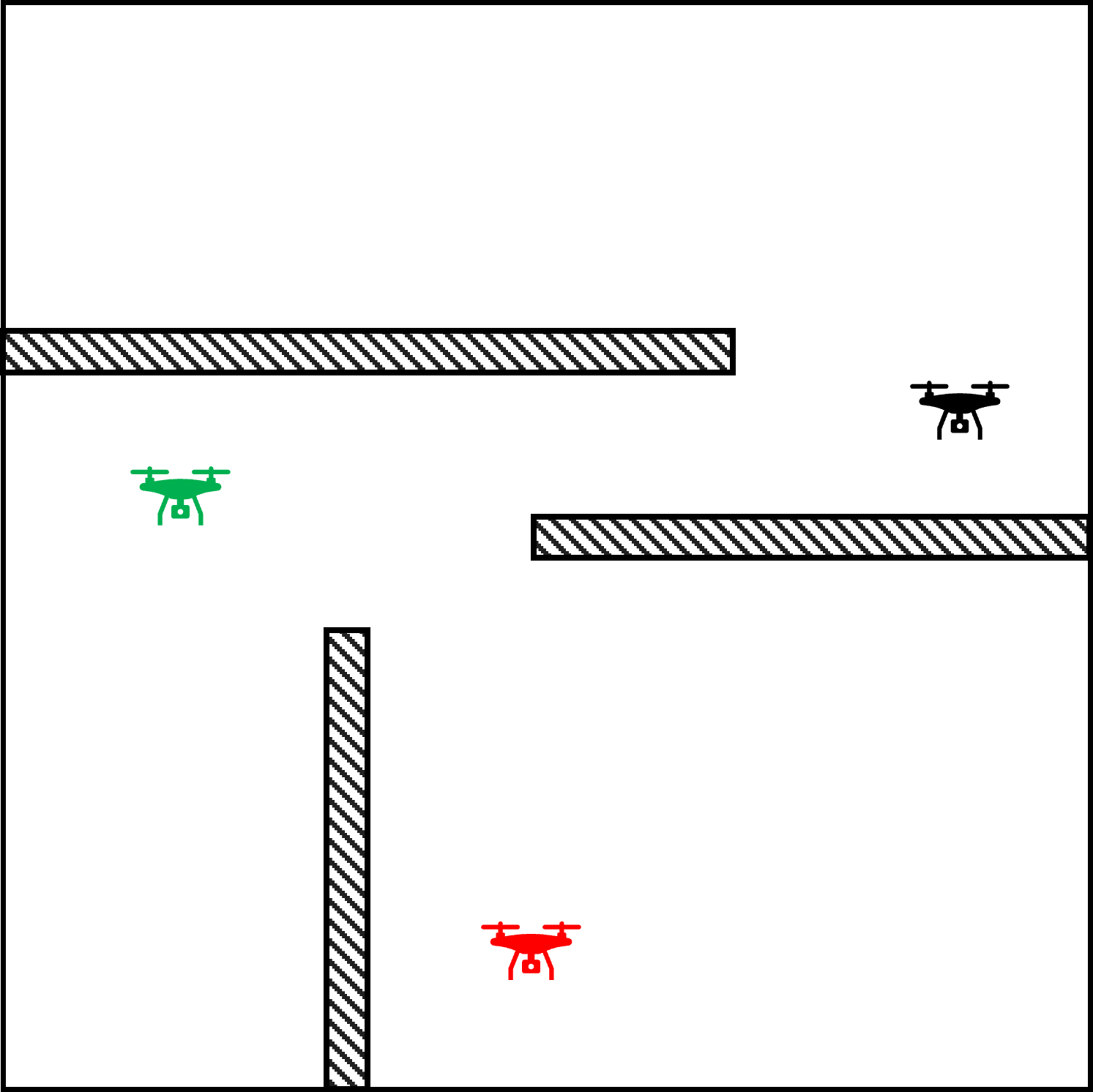}
         \caption{}
         \label{FigScenario3}
     \end{subfigure}
     \caption{Three examples showing the different scenarios to be addressed by the agents, including cases of (a) complex environments with obstacles, (b) unreachable targets, and (c) no targets due to false alarms.}
        \label{FigScenarios}
\end{figure}

Figure \ref{FigScenarios} shows representations of the different possible scenarios for the target localization problem in environments with uncertainties. In realistic environments, the agents might encounter several obstacles that hinder their movements and affect their decision-making and planning (Fig. \ref{FigScenario1}). In addition to obstructing the movement of the agents, obstacles also attenuate the sensor readings, making it more difficult to detect signals emitted from the target. As a result, an agent with a higher reading might have to travel a larger distance to reach the target when compared to an agent with a lower reading. For example, the black agent in Fig. \ref{FigScenario1} has a higher sensor reading than the red agent (since it is closer to the target), but the red agent needs less time to reach the target. Such cases increase the difficulty of the search process and require more cooperation and planning from the agents. Moreover, since target localization is common in applications with harsh environments (i.e. search and rescue and fire localization), there could be cases where the target can be detected but is unreachable (Fig. \ref{FigScenario2}). In such cases, the agents are required to cooperate to determine that the target exists but is unreachable and then provide an estimate of the target location. Finally, there are cases where the agents are deployed with uncertainty about the target's existence (Fig. \ref{FigScenario3}). Here, instead of blindly searching the entire AoI, the agents should cooperate to intelligently determine whether a target exists or not with minimum resource consumption.  

In all scenarios, it is desired for the task to be executed as quickly as possible and with low resource consumption, which requires the agents to cooperate. Although the target location is unknown, the agents can use their observations to quickly localize the target. Some of these observations, such as sensor readings, can guide the agents towards areas with high probabilities of containing the target. Other observations, such as the distribution of other agents and the layout of the environment, can help the agents coordinate and optimize their collective search paths as well as their resource consumption. In this context, agents can cooperate to expedite localization or to conserve resources. Resource conservation can be achieved if agents assess their contributions to the task and remain idle when they are not needed. For the scenario in Fig. \ref{FigScenario1} as an example, given the layout of the environment and that the red and black agents have informative sensor readings, the green agent should intelligently decide to maintain an idle state while the other agents search the environment, since it cannot contribute to the task. In cases where all the agents do not have informative readings, they need to cooperate and spread to better search the environment to collect more readings or reach conclusions about the existence and reachability of the target.

\section{Related Work}
\label{Sec: RelatedWorks}

The initial works addressing the target localization problem focused on reducing the target location estimation error through advanced mathematical data fusion techniques or through the efficient placement of sensing nodes. The authors in \cite{bai2014maximum, chin2008accurate} use methods such as Inverse Square Law and Maximum Likelihood Estimation (MLE) to fuse the data readings collected and estimate the location of the target. In \cite{pandey2017event, grondin2019sound}, the authors use Direction of Arrival (DoA) and Time Difference of Arrival (TDoA) for sound localization. In \cite{liu2010analysis}, the authors use Bayesian Methods to iteratively fuse data over time and build and update a probability distribution over the possible locations of the target. Alternatively, several methods focus on the deployment of sensing nodes to enhance localization accuracy and time. In \cite{liu2010analysis, liu2017node}, the authors proposed greedy- and genetic-based optimization methods to optimize the deployment of sensing nodes in an area for efficient target localization. Rather than focusing on the deployment of sensing nodes, several works considered the problem of active node selection, where many nodes are already deployed, and the system aims to optimize resource expenditure through intelligently activating informative nodes. The authors in \cite{alagha2019data, alagha2020rfls, alagha2021sdrs, alagha2022influence} introduced data-driven methods for the selection and activation of nodes, considering factors such as area coverage and data confidence. In \cite{shurrab2022iot}, the authors propose an RL-based method for active node selection using Q-Learning. While the aforementioned proposals prove efficient in estimating the target location and managing resources, they struggle with adapting to the dynamicity of the varying environments, and they lack intelligent decision-making in the absence of data readings.

Alternative and relatively more recent methods to address the target localization problem rely on mobile sensing agents that survey the area to find the target. Initial methods use pre-defined survey paths, such as the uniform search method \cite{ziock2002lost} which is common in many works due to its simplicity. In \cite{lazna2018cooperation}, the authors propose circular path planning strategies using the directional characteristics of sensors. The authors in \cite{xiao2017sampling} propose area surveying method based on a data-driven approach that uses Bayesian methods to decide what actions to take next based on probability density functions (PDFs) about the target location. These methods prove difficult to adapt to complex environments due to the lack of intelligent cooperation between the sensing agents, as well as the need for re-modeling for varying environments. Alternatively, recent methods explored the use of RL to address this issue. In \cite{liu2019double}, the authors train an agent to search for a target in a complex environment with obstacles, using Double Q-Learning with CNNs. Here, the agent's observations are represented as a stack of 2D maps capturing information about the environment and are given to the CNN for decision-making. The work proves efficient in tackling the problem but has drawbacks in terms of scalability when extrapolated to multi-agent settings. More recent works \cite{alagha2022target, alagha2023multi} approached the problem in multi-agent settings using MADRL by developing methods that motivate the agents to collaborate to efficiently and quickly find the target in complex environments. However, these works have the assumption that the target always exists and is always reachable, which does not account for scenarios of false alarms or the obstruction of the path to the target. This requires a more complicated decision-making process where the agents need to make more than just mobility-based decisions to find the target.

\section{Proposed System}
\label{Sec: PropSystem}

\begin{figure}[h]
    \centering
    \includegraphics[width=0.6\columnwidth]{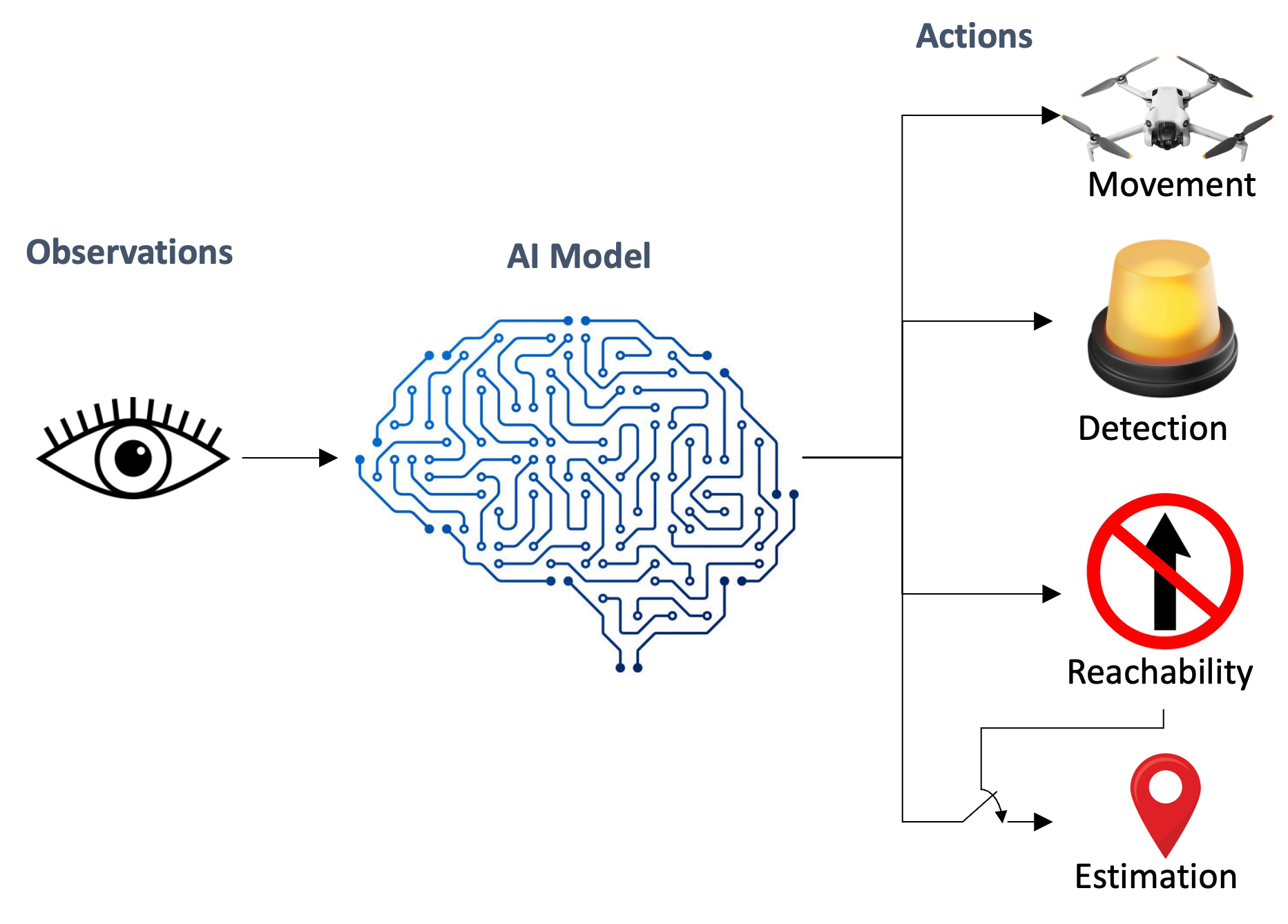}
    \caption{An overview of the model proposed, which is to be deployed on each sensing agent.}
    \label{GeneralOverviewFig}
\end{figure}

This section presents the MADRL method used to intelligently produce sensing agents capable of tackling the different complex scenarios of target localization. An overview of the final proposed model is presented in Fig. \ref{GeneralOverviewFig}. At each timestep, and for a given agent, the MADRL model translates its observations into one of three possible action dimensionalities: Movement, Detection, and Reachability. Movement actions are responsible for controlling the mobility of the sensing agent in the environment. A detection action is concerned with flagging the existence of a target in the AoI, with the aim of conserving resources if the target does not exist. A reachability action determines if a target exists but is unreachable due to environment complexity. Given that the target is unreachable, an estimation process is triggered, which is responsible for estimating the location of the unreachable target. The key challenge here is to produce all actions, as well as the estimation process, in one AI model capable of translating an agent's observations, while ensuring its cooperation with other sensing agents. This section formulates the MADRL problem and discusses the modeling of the MADRL methods used to obtain the final model.

\subsection{MADRL Formulation and Policy Optimization}
\label{subsec: MADRL Formulation}
In the context of MADRL, Markov Games are generally used to extend Markov Decision Processes (MDPs) into multi-agent settings \cite{gronauer2021multi, alagha2023blockchain, alagha2024blockchain}. In the problem of target localization, the state at a given instant is defined by the distribution of agents and obstacles, as well as the target location. Given that the location of the target is unknown, the problem is represented as a Partially Observable Markov Game (POMG). The different components of a POMG include a set of $\mathcal{S}$ finite states, finite action sets $\mathcal{A}_1 , \mathcal{A}_2, ... , \mathcal{A}_N$ for each of the $\mathcal{N}$ agents, finite observation sets $\mathcal{O}_1 , \mathcal{O}_2 , ... , \mathcal{O}_N$, a state transition function $\mathcal{P}(s',\textbf{o}$ $|$ $s,\textbf{a})$ that computes the probability of ending up in state $s'$ with observation $\textbf{o}$ after taking action $\textbf{a}$ in state $s$, a reward function $\mathcal{R} : \mathcal{S} \times \mathcal{A} \rightarrow \varmathbb{R}$, and a discount factor $\gamma \in [0,1]$. Here, $\textbf{a} = (a_1,...,a_N)$ and $\textbf{o} = (o_1,...,o_N)$ denote joint actions and observations from the $\mathcal{N}$ agents at a given instant. In MADRL settings, the target localization task occurs in discrete steps. During each step, agent $i$ uses its policy $\pi_i : \mathcal{O}_i\times\mathcal{A}_i \rightarrow [0,1]$ to translate an observation $o_i \in \mathcal{O}_i$ into an action $a_i \in \mathcal{A}_i$, and receives a reward $r_i$. The primary aim of each agent is to maximize the cumulative rewards earned throughout the duration of an episode.

Using the collected experiences, PPO \cite{schulman2017proximal} is utilized to update the decision-making policies of the agents. PPO is a policy gradient (PG) method that has two components, an actor and a critic. The actor (policy) network takes the current observations as input and produces a probability distribution over the possible actions. The critic (value function) predicts the future rewards, which is used in updating the actor network. The objective is to optimize the actor policy $\pi_\theta$, parametrized by $\theta$, to maximize the cumulative rewards in an episode. PPO strikes a balance between simplicity and performance by using a clipped surrogate objective that ensures stable and efficient policy updates, which is given as:

\vspace{-0.5em}
\begin{equation}
    L^{CLIP}(\theta) = \hat{\varmathbb{E}}_t \left[\text{min}(r_t(\theta)\hat{A}_t, \text{clip}(r_t(\theta), 1-\varepsilon, 1+\varepsilon)\hat{A}_t) \right]
    \label{eq. clipped surr}
\end{equation}

where $\varepsilon$ is a hyperparameter that controls the clipping. On the other hand, $r_t(\theta) = \frac{\pi_\theta(a_t|s_t)}{\pi_{\theta_\text{old}}(a_t|s_t)}$ is a probability ratio between the old and the current policy of taking a given action. The advantage function estimate ($\hat{A}_t$) quantifies how good taking a specific action $a_t$ in state $s_t$ is. Here, $\hat{A}_t$ is estimated using Generalized Advantage Estimate (GAE) \cite{schulman2015high}. After collecting $H$ steps of experience in the environment (horizon length), PPO is used to update the policy, where the estimator of $\hat{A}_t$ is computed as:

\vspace{-0.3em}
\begin{equation}
    \label{eq:advantage}
    \hat{A}_t^{\text{GAE}(\gamma, \lambda)} = \sum_{l=0}^{H} (\gamma \lambda)^{l} \delta_{t+l}, \quad \delta_{t+l} := r_{t+l} + \gamma V(s_{t+l+1}) - V(s_{t+l})
\end{equation}
where $\delta_{t+l}$ is the Temporal Difference (TD) error, $\gamma \in [0,1]$ is the discount factor that controls the weight of future rewards, $\lambda \in [0,1]$ is a parameter that controls the bias-variance trad-off, and $V(s_t)$ is the value function estimate. 

As per the original work in \cite{schulman2017proximal}, the policy surrogate is further improved by considering two additional terms, namely $L_t^{VF}(\theta)$ and $S[\pi_\theta](s_t)$, which results in:

\vspace{-0.3em}
\begin{equation}
    L^{CLIP+VF+S}(\theta) = \hat{\varmathbb{E}}_t \left[L_t^{CLIP}(\theta) - c_1 L_t^{VF}(\theta) + c_2S[\pi_\theta](s_t) \right]
    \label{FinalObjectiveFunction}
\end{equation}
where $c_1$ and $c_2$ are coefficients. Here, $L_t^{VF} = (V_\theta(s_t)-V_t^{\text{targ}})^2$ is the squared-error loss that ensures the value function is accurately approximated, while $S$ encourages exploration by increasing the entropy of the policy.

\subsection{Observation Space}
\label{Sec: observation space}
\begin{figure}[h]
    \centering
    \includegraphics[width=\textwidth]{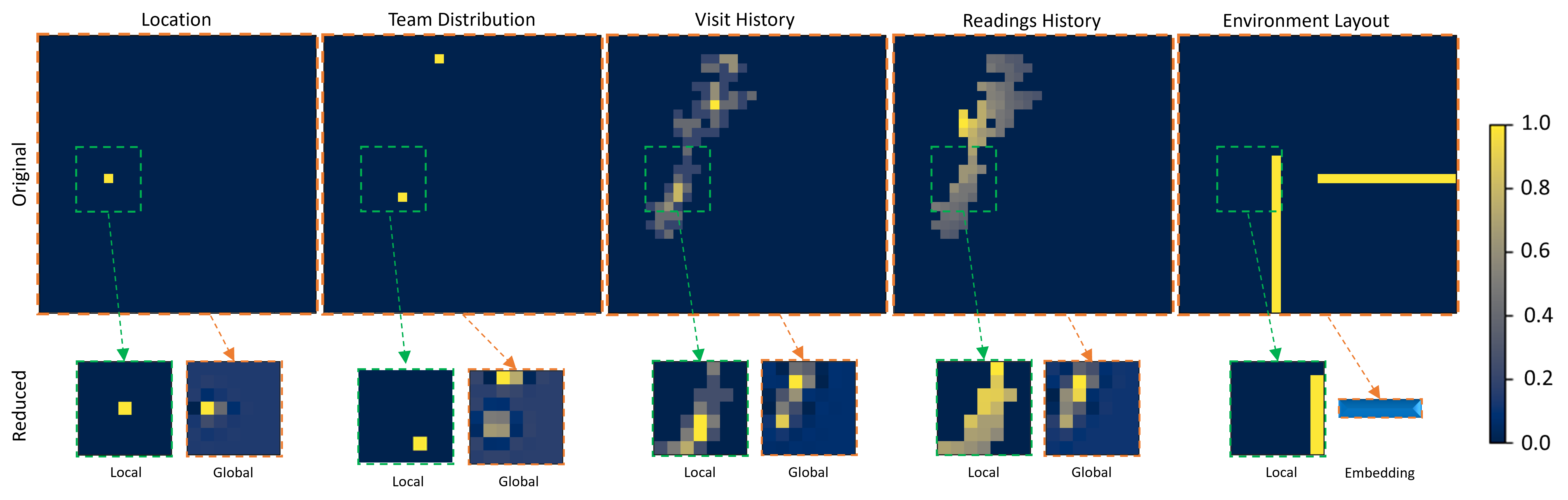}
    \caption{The five collected observations by a sensing agent in a team of three agents. The original observations (top row) are processed to obtain the reduced observations (bottom row). The reduced observations are either local (green) or global (orange).}
    \label{FigObservations}
\end{figure}

An efficient decision-making process in target localization depends on the collected sensor readings, the distribution of sensing agents and the areas they visited, and the layout of the environment and its complexities. For efficient execution of the given task, all such observations need to be tracked and stored over time. To tackle this, the observations of each agent are represented as 2D maps that capture both current and past spatial information collected throughout the task. These maps encapsulate information about the agents' locations, the collected readings, previously visited areas, and the layout of the environment, which are needed for the decision-making process. To achieve this, the AoI is regarded as an $h\times w$ grid, and the observations are represented as stacks of $h\times w$ maps showing spatial information throughout the AoI. In this work, at a given timestep $t$, the agent collects information and updates 5 observations, which are shown in Fig. \ref{FigObservations}. Among these observations, the \textit{Location} observation indicates the location of the agent in the AoI. The \textit{Team Distribution} observation highlights the locations of other agents in the team with respect to the AoI. The \textit{Visit History} observation records the visiting frequency of the grid elements by the team members. The \textit{Readings History} observation tracks the readings collected and their spatial pattern in the AoI. The \textit{Environment Layout} observation highlights the different obstacles in the environment. All of those observations, given as 2D maps, undergo normalization before going to the agent's policy which ensures learning stability and faster convergence \cite{sola1997importance}.

It is worth mentioning that due to temporal aspect of the problem, using methods in Recurrent Neural Networks (RNNS), such as Gated Recurrent Units (GRUs) and Long Short-Term Memory (LSTM), are popular in DRL applications. This is because such methods efficiently capture the temporal aspects of the decision-making process, allowing agents to correlate long sequences of actions. However, target localization problems are also spatial in nature, which is an added complexity when using RNNs. Instead, the proposed approach accumulates historical data in 2D maps to remove the temporal dependency. For a given timestep, the agent has access to its previously collected and stored data, in addition to new data collected in the current step. This allows the agent to act solely based on its current observation (that already includes previously accumulated data), which transforms the problem into a less complex one that is focused purely on spatial correlation leveraging simple CNNs. This is also common in existing literature as shown in \cite{liu2019double, damani2021primal}. Here, the collected information could be accumulated over time throughout the task by just updating each observation without changing its dimensionality. Target localization tasks rely heavily on the spatial correlation of sensor readings and agents' distribution, which can only be captured in 2D maps. The aforementioned observations are meant to guide the agents towards efficient target localization. Ideally, the readings history map should guide the agent in the direction of higher data readings. At each step, the agents update this map with the latest sensor readings. However, in many cases, the agents might not have sufficient readings (because of being distant from the target for instance) and hence require more exploration of the environment. Here, the location, team distribution, and visit history maps help the agents coordinate to explore the environment efficiently. Knowing the team distribution also helps an agent preserve resources when possible if it is evident that the other agents are getting closer to the target and that the agent can no longer be of benefit to the search task. The environment layout map is essential in translating all the aforementioned observations since it significantly affects the readings collected as well as the environment exploration planning process. In this work, one assumption is that the agents can communicate and share information to be used in updating the aforementioned maps.

To further optimize the learning and decision-making processes and reduce their complexity, the five observations undergo further pre-processing steps. Here, the aim is to reduce the dimensionality of the observations while maintaining their informativeness. To this end, as seen in Fig. \ref{FigObservations}, the five original observations are reduced into ten optimized ones, which are split into global and local observations. The agent can use the local observations to take actions based on its surroundings, such as following higher readings or avoiding obstacles. Global observations summarize the environment's spatial features, which help in making coordination and planning decisions. Out of the reduced observations, the first 9 have a dimension of $n \times n$, where $h, w > n > 1$ and $n$ is odd. Here, $n$ is considered a tunable hyperparameter, where smaller values entail lower computational complexity and higher information loss, when compared to higher values. At timestep $t$, local observations are acquired by capturing a $n \times n$ window that is centered at the agent's position from the corresponding original observation, capturing surrounding information. Global observations are computed by downsampling the original observations using bi-linear interpolation. For the environment layout, the use of generic downsampling proved inefficient due to significant loss of information, particularly with the increasing complexity of the environment with more obstacles and narrow openings. To circumvent this issue, we create embeddings that encapsulate essential information from the environment at a lower dimensionality using Convolutional AutoEncoders (CAE). A CAE is composed of an encoder and a decoder. An encoder is a CNN responsible for converting the input map into a 1D vector (embedding) through several convolution and fully connected layers. The decoder is responsible for attempting to generate the original input map starting from the embedding. The decoder's job is to assess the power of the encoder in creating embeddings that properly represent the original input at a lower dimensionality. In this work the encoder consists of 3 convolutional and 2 fully connected (FC) layers, where the last FC layer represents the embedding. The decoder takes the embedding as an input into 2 FC layers followed by 2 convolutional layers that produce the reconstructed map. A synthetic dataset of different environment layouts is used to train the CAE before the MADRL training. The training of the CAE aims to minimize the reconstruction error between the input and the reconstructed map. After the training is completed, the encoder can generate embeddings for the environment layout, while the decoder is discarded.


\subsection{Action Space}
\label{subsec: Action Space}
To address the complex scenarios of target localization discussed in Section \ref{sec: Problem Formulation}, the action space at a given step for an agent is divided into mobility, detection, and reachability actions. In terms of mobility, the agent has a fixed speed and decides on the direction of movement. Given $D$ possible discrete directions \{1, 2, ..., $d_i$, ..., $D$\}, the movement angle is given as:

\begin{equation}
\theta = 2\pi \frac{d_i}{D}
\end{equation}
where the hyperparameter $D$ determines how detailed the movement is. In this work, we use $D=8$, which allows the agent to move in one of the cardinal or ordinal directions. Assuming a fixed speed, the agent can move a fixed distance in one of these directions with the aim of contributing to the localization task. It has been found in this work that discretizing the direction of movement into 8 possible values is sufficient. On the other hand, choosing to stay idle, if needed, helps in preserving resources.

The detection and reachability actions are each represented by a binary value to flag the existence of the target and its reachability, respectively. At a given time step $t$, an agent could determine that the target does not exist and hence set the existence flag to 1. Similarly, an agent could determine that a target is unreachable by setting the reachability flag to 1.

The decision-making process is discretized, where at each step, an agent can choose only one of the 11 actions. If the majority of the agents declare that the target does not exist, the search process stops. Similarly, if the majority declare that the target is unreachable, a target estimation process is triggered, which estimates the target location (to be explained later in Section \ref{Sec: Target Estimation}). In certain time steps, some actions may not be possible, such as moving outside the boundaries of the AoI or into a wall. Such invalid actions are masked out during the decision-making process, where the agent only chooses from the available actions.

\subsection{Policy Networks and Learning Process}
\label{Sec: Policy and Learning}

As discussed in Section \ref{subsec: MADRL Formulation}, we use PPO to train the MADRL agents. Generally, the actor and critic in PPO are represented by Deep Neural Networks (DNNs). In this work, CNNs are used for the actor and critic, because the observations of each agent are represented as 2D maps. CNNs are crucial for the target localization problem as they effectively correlate spatial features within the input maps.

\begin{figure}[h]
    \centering
    \includegraphics[width=\textwidth]{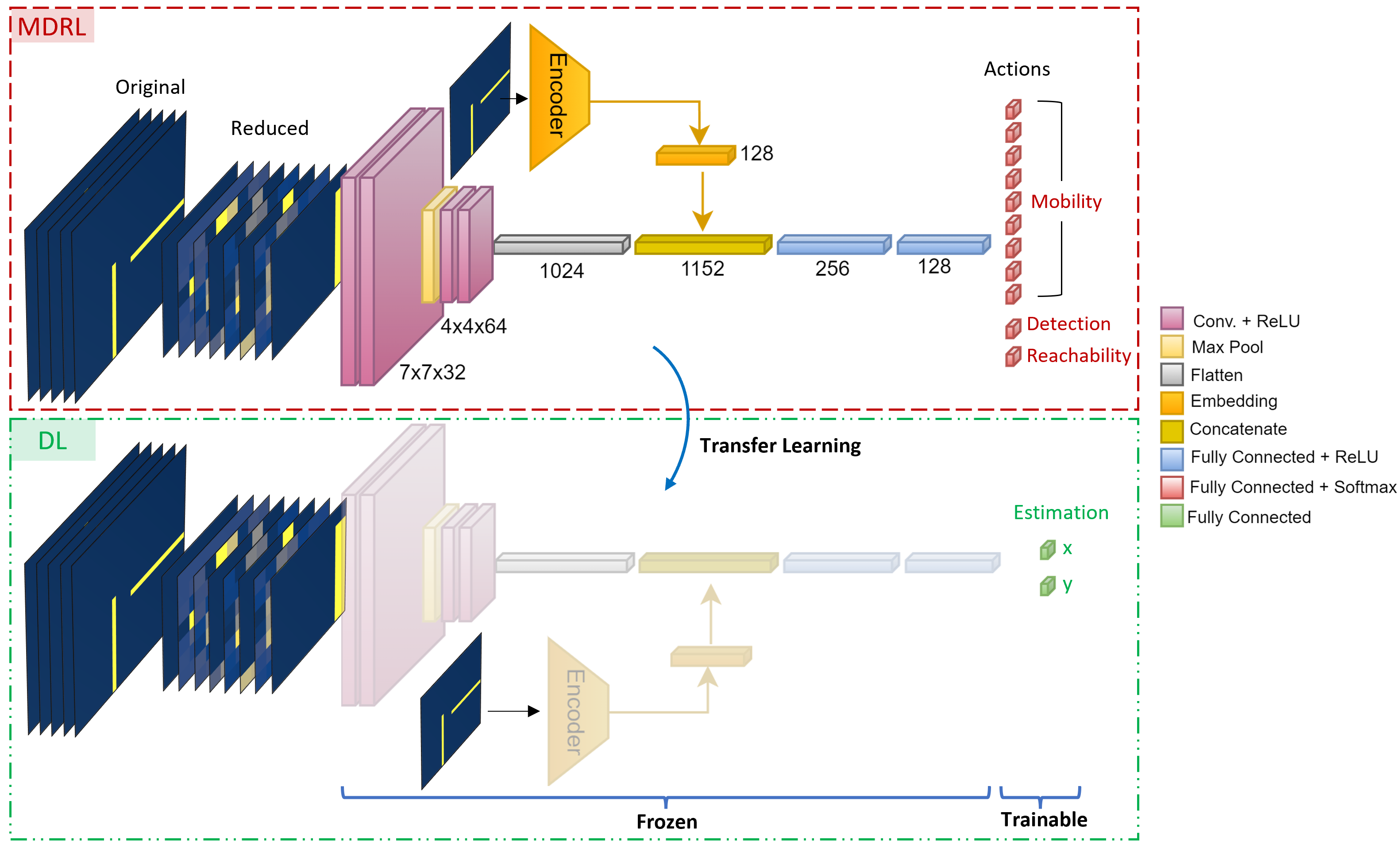}
    \caption{The actor architecture trained through MADRL (top), and the architecture of the estimation model trained through DL and TL (bottom).}
    \label{FigTraininigProcess}
\end{figure}

Fig. \ref{FigTraininigProcess} shows the architecture used for the actor (upper part of the figure), which is similar to the LeNet-5 architecture \cite{lecun2015lenet}. The network takes the first 9 reduced observations as input, which are then processed in convolution and max pooling layers for feature extraction. The embedding for the environment layout observation is concatenated with the processed and flattened observations, before being fed into the fully connected layers, . The fully connected layers produce 11 outputs, which are fed into a softmax function that produces a probability distribution for the possible actions. During the decision-making process, the agent samples an action from this distribution, which is then executed in the environment.

The proposed MADRL method uses PPO in multi-agent settings, sometimes referred to as Multi-Agent PPO (MAPPO), with Centralized Learning and Distributed Execution (CLDE) \cite{baker2019emergent}. Here, a copy of the actor is given to each agent, where agents act independently based on their observations. During the training stage, a centralized critic is used to assess the experiences collected by the agents. The architecture of the critic is similar to that of the actor, but with 1 output representing the value function. Using a centralized critic is essential in addressing the non-stationarity problem that is common in MADRL due to the influence agents have on each other's view of the environment \cite{lyu2021contrasting}. The value function, which is the critic's output, is used during the PPO process to update the actor and the critic.

A \textit{team-based shaped} reward function is proposed in this work to guide the learning. A team-based (joint) reward function gives equal rewards to the agents according to the collective behavior. As a result, the agents would be motivated to act in a way that benefits the entire team. A shaped reward gives more frequent feedback to the agents during the task, which helps speed up the learning process. Following the joint action taken by the agents in a given time step, the environment returns to all the agents a similar reward. To achieve all the desired behaviors, the reward function after step $t$ is given as:

\begin{equation}
    R_t = \begin{cases}
             - Q & \text{if flags are incorrect} \\
             - v + 1  & \text{if } \text{min}(D_t) < \text{min}(D_{t-1})\\
             - v - 1 & \text{otherwise}
       \end{cases}
    \label{eq:rewardequation}
\end{equation}
where Q is a large penalty, $v$ is the total number of mobility actions taken, and $D_t$ represents the set of distances between the agents and the target. In the function, the first condition gives the team a high negative reward (in this work $Q=500$) if a target is incorrectly declared non-existent or unreachable. Alternatively, in the second and third conditions, the agents are rewarded according to their proximity to the target and their resource consumption. At each timestep, the agents are penalized by ($-v$), where $v$ is the number of agents who moved in the environment. This indicates resource consumption, which is essential in pushing the agents towards finishing the task as fast as possible. It also motivates the agents to only move in the environment if needed and stay idle otherwise. The agents receive an additional positive or negative reward ($\pm 1$) based on their proximity to the target. The reward would be positive if they have moved closer to the target from step $t-1$ to $t$, otherwise they receive a negative reward. The team is considered to have moved closer to the target if the nearest agent(s) at step $t-1$ took a mobility action towards the target at step $t$, meaning min($D_t$) $<$ min($D_{t-1}$). Due to the environment complexity and the existence of obstacles, it is inaccurate to compute the travel distance between an agent and the target using generic methods like Manhattan and Euclidean distance. Instead, Breadth First Search (BFS) is used, which determines reward calculations. BFS explores different paths starting from an initial node until the goal is reached, aiming to find the shortest path. It is worth mentioning that the agents have no knowledge of the location of the target, and act only based on their observations. The reward function serves as a feedback mechanism, used only during the training stage, to improve the policies of the agents using PPO.

The training process is done using a CLDE method \cite{baker2019emergent}, in which the MADRL model is optimized is based on PPO. In this method, the agents act in a distributed manner in the environment according to their actor network copies. The experiences collected by the agents are then gathered and used centrally for updating the actor and critic. Once the training is over (which usually occurs in simulations), the final model is deployed on the sensing agents which act independently based on their observations. Algorithm 1 explains the training process in CLDE-based PPO. The learning takes place over episodes of the problem, where the environment is reset to a different initial state at the beginning of each episode. During episode step $i$, each agent $j$ samples an action $a_j^i$ by feeding its observations $o_j^i$ to its actor network. The agents then step in the environment with their actions, which then returns a reward value $r^i$, a new set of observations $\textbf(o)^{i+1}$, and a termination flag $d^i$ indicating if the episode has ended.  When the number of training timesteps reaches the horizon ($H$), the actor and critic networks are updated using the gathered experiences. The process repeats for a pre-defined total number of time steps.

\begin{table}[!ht]
\setlength{\tabcolsep}{3pt}

\centering
\begin{tabular}{p{240pt}}
\hline
\textbf{Algorithm 1:} The training process of CLDE PPO\\ 
\hline
\textbf{Input}: Initialized actor and critic \\
1: \textbf{for} Step $\in$ TotalSteps \textbf{do}:\\ 
2: \hspace{0.25cm} $\textbf{o}^0$ = ResetEnvironment()\\
3: \hspace{0.25cm} \textbf{for} $i$ = 0, 1, 2, ..., MaxEpisodeLength:\\
4: \hspace{0.5cm} \textbf{for} $j$ = 1, 2, ..., $\mathcal{N}$:\\
5: \hspace{0.75cm} $a^i_j$ = Sample(Actor($o^i_j$))\\
6: \hspace{0.5cm} \textbf{end for}\\
7: \hspace{0.50cm} $\textbf{a}^i$ = [$a^i_1, a^i_2, ...$]\\
8: \hspace{0.50cm} $\textbf{o}^{i+1}, r^i, d^i$ = Step($\textbf{a}^i$) \\
9: \hspace{0.50cm} Save Experiences\\
10: \hspace{0.35cm} \textbf{if} Step \% $H$ == 0 \textbf{then}:\\
11: \hspace{0.6cm} Update\_PPO()\\
12: \hspace{0.35cm} \textbf{end if}\\
13: \hspace{0.35cm} \textbf{if} $d^i == 1$ \textbf{then} break \\
14: \hspace{0.1cm} \textbf{end for}\\
15: \textbf{end for}\\
\hline

\end{tabular}
\label{First Model training}
\end{table}

\subsection{Target Estimation with Transfer Learning}
\label{Sec: Target Estimation}
\begin{figure}[h]
    \centering
    \includegraphics[width=0.9\columnwidth]{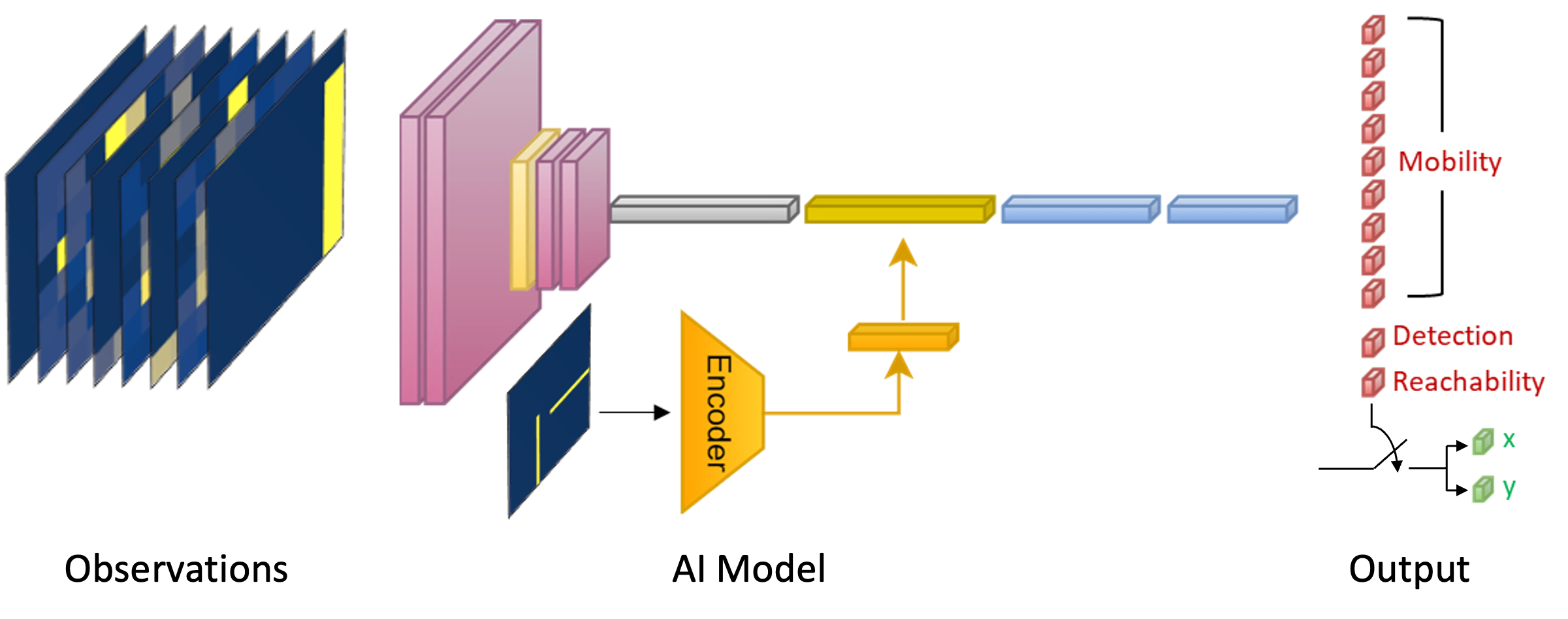}
    \caption{The final model deployed on each of the sensing agents.}
    \label{FigFinalModel}
\end{figure}

The MADRL model discussed in Section \ref{Sec: Policy and Learning} is responsible for regulating the continuous decision-making process for each agent throughout the localization task. However, if a target is determined unreachable, it is essential to provide an estimate for its location. To reduce the complexities of the MADRL training process and the final model, the target estimation is not performed at each time step, but only when target unreachability is determined. To achieve this, a separate model is trained using a typical DL process with TL. Here, based on the observations collected until unreachability is determined, the aim is to provide an estimate for the (x,y) coordinates of the target. A dataset is pre-built using sets of observations and the corresponding target locations to be estimated, which are used to train the DL model to estimate the location based on the observations. To reduce the training and deployment complexities, we use TL to utilize knowledge from the previously obtained MADRL model, as shown in Fig. \ref{FigTraininigProcess}. Here, rather than training a new DL model from scratch for target estimation, the weights of the MADRL actor network are initially copied into the new DL model. During the training process, these layers are frozen, i.e. not trained, and only the last fully connected layer is trained. This is viable because the initial layers, especially the convolution layers, have already been trained to extract features related to target localization in the MADRL process. Such features would still be similar in the target estimation model, and hence only a final classification layer would be sufficient. Following the training process, and since both the MADRL and DL models have common initial layers, they can be combined into one model with two output heads; one for continuous decision making (mobility, detection, and reachability) and one for target estimation, as shown in Fig \ref{FigFinalModel}. Both action heads take the same set of extracted features from the initial layers, but only differ in the weights of final layer, with the target estimation output head only getting triggered when unreachability is determined.

It is worth mentioning that TL is used solely during the training stage, which generally occurs offline before deployment, managed by the system administrator. Hence, there are no privacy concerns during the training when it comes to transferring knowledge from the MADRL model to the estimation model. In case this work is extrapolated such that the training itself happens in a distributed manner with knowledge exchange between the agents, existing methods for privacy preservation can be used, such as differential privacy for TL in MADRL \cite{cheng2022multi, shen2024privacy}.

\section{Simulation and Evaluation}
\label{sec: simulations}

Extensive experiments are conducted in this section to validate the efficiency of the proposed method, as well as benchmark it against existing works in the literature. All the simulations have been conducted using an Intel E5-2650 v4 Broadwell workstation equipped with a 128 GB RAM, an 800 GB SSD, and an NVIDIA P100 Pascal GPU (16 GB HBM2 memory).

\subsection{Simulation Environment}
\label{subsection: Simulation Environment}
To validate the proposed methods, a sample environment of radioactive source localization is used, where the agents are tasked with localizing a radiation source in an area of size 1km $\times$ 1km. The environment is simulated using radiation physics \cite{liu2010analysis, knoll2010radiation}, where the readings collected by the detectors carried by each agent follow a Poisson distribution. Given a certain source strength $S$ and a distance $d$ from the radiation source to the detector, the photon counts per minute ($CPM$) are given as:

\begin{equation}\label{CPMEquation}
CPM \propto \frac{S}{d^2}
\end{equation}

This essentially indicates that radiation sources with higher strength can be better detected from a longer distance. Due to the existence of obstacles, the $CPM$ collected by a certain detector is influenced based on the number of obstacles. Generally, each obstacle has an attenuation factor that represents the absorption/reflection of photons. In this work, we assume constant attenuation of $\mu = 0.1$ for each obstacle. While the proposed methods are tested on a radiation localization environment, they are still valid for any localization environment. This is mainly because data collected in localization problems are always a function of the distance between the target and the agent; the closer the agent is the higher the readings, based on the Inverse Square Law. Several applications are based on these representations, including radiation, sound, and heat localization \cite{chin2008accurate, davis2014sound}.

For all the following experiments, each model is trained for 30 million environment steps using MADRL. At the beginning of each episode, the target location, the agents' initial distribution, and the environment layout (with varying number of obstacles) are randomized. Hence, the model is trained on variations of random combinations of the possible states. Additionally, each episode is randomly set to one of three possible scenarios: 1) target exists and reachable, 2) target exists and unreachable, and 3) target does not exist, where each scenario has an equal chance of occurring. An episode terminates if the agents make the right decisions based on the scenario, or if a 100-timestep limit is reached. Following each 40k training steps, the agents are placed in a testing environment for 4k steps where they follow the most probable actions based on their policy (instead of sampling). The average performance of the testing steps is then recorded and plotted. The set of hyperparameters used in the PPO and CAE methods is shown in Table \ref{Hyperparameters}. The PPO hyperparameters are based on the original work in \cite{schulman2017proximal}.

\begin{table}[ht]
\caption{PPO and CAE Hyperparameters}
\vspace{-1em}
\setlength{\tabcolsep}{0pt}
\begin{center}
\begin{tabular}{|P{0.25\columnwidth}|P{0.25\columnwidth}|}
\hline
PPO & Value\\
\hline
 $\varepsilon$ & $0.2$\\
 $\gamma$ & $0.99$\\
 $\lambda$ & $0.95$\\
 $H$ & $4000$\\
Epochs per update & $20$\\
Learning rate & $3\times 10^{-4}$\\
\hline
CAE & Value\\
\hline
Learning rate & $1\times 10^{-3}$\\
Embedding Size $d$ & $128$\\
Dataset Size & $500000$\\
\hline
\end{tabular}

\end{center}
\label{Hyperparameters}
\vspace{-2em}
\end{table}

\subsection{MADRL Performance Analysis}
\label{subsection: Performance}

The performance of the proposed MADRL model is analyzed in terms of the collected reward, episode length, and cost. The episodic reward reflects the total accumulated reward collected in an episode, on average. The episodic length represents the number of steps until an episode is terminated, which reflects how quick the agents are in finishing the task. The episodic cost reflects the total number of movement actions taken during an episode, which reflects resources consumption. The lowest movement cost in a given time step is 0 if none of the agents moved, while the highest is $N$ if all the agents have moved.  

\begin{figure}[h]
     \centering
     \begin{subfigure}{0.32\columnwidth}
         \centering
         \includegraphics[width=\linewidth]{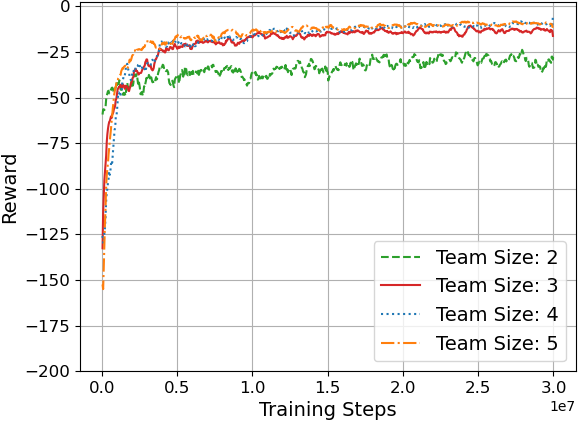}
         \caption{Target Strength: $1\times 10^9$}
         \label{ResultsReward1}
     \end{subfigure}
     \begin{subfigure}{0.32\columnwidth}
         \centering
         \includegraphics[width=\linewidth]{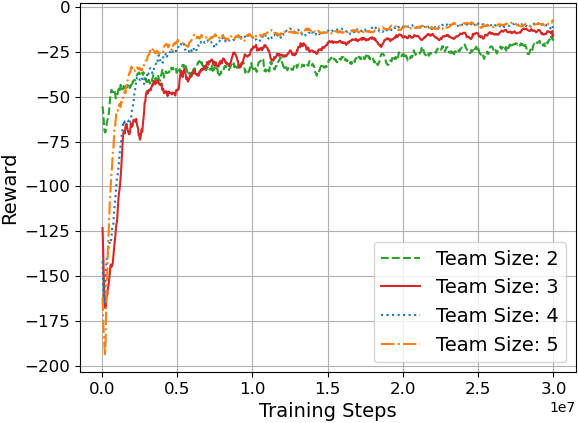}
         \caption{Target Strength: $5\times 10^8$}
         \label{ResultsReward2}
     \end{subfigure}
     \begin{subfigure}{0.32\columnwidth}
         \centering
         \includegraphics[width=\linewidth]{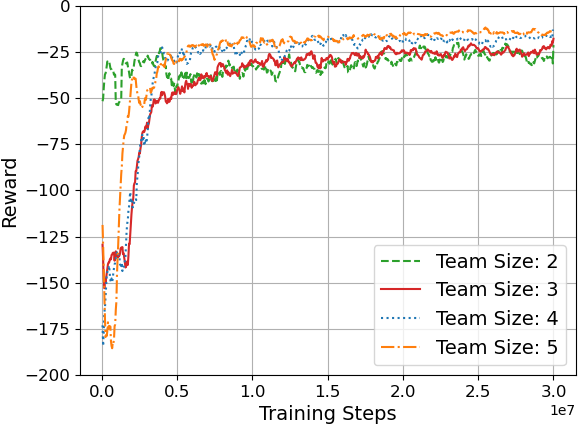}
         \caption{Target Strength: $1\times 10^8$}
         \label{ResultsReward3}
     \end{subfigure}

     \hfill
     \hfill

     \vspace{-0.5em}

    \begin{subfigure}{0.32\columnwidth}
         \centering
         \includegraphics[width=\linewidth]{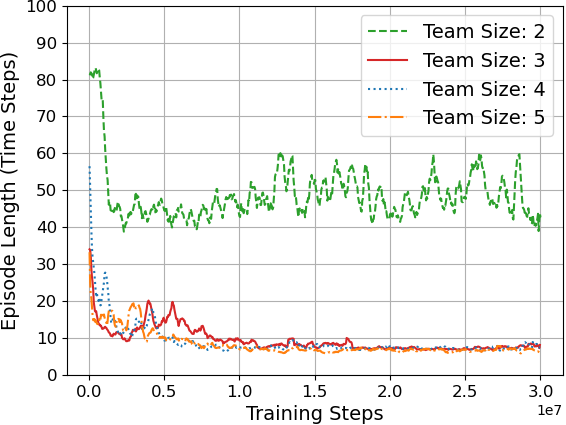}
         \caption{Target Strength: $1\times 10^9$}
         \label{ResultsTime1}
     \end{subfigure}
     \begin{subfigure}{0.32\columnwidth}
         \centering
         \includegraphics[width=\linewidth]{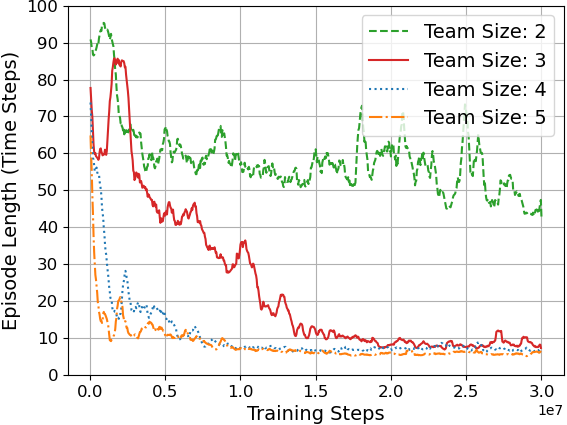}
         \caption{Target Strength: $5\times 10^8$}
         \label{ResultsTime2}
     \end{subfigure}
     \begin{subfigure}{0.32\columnwidth}
         \centering
         \includegraphics[width=\linewidth]{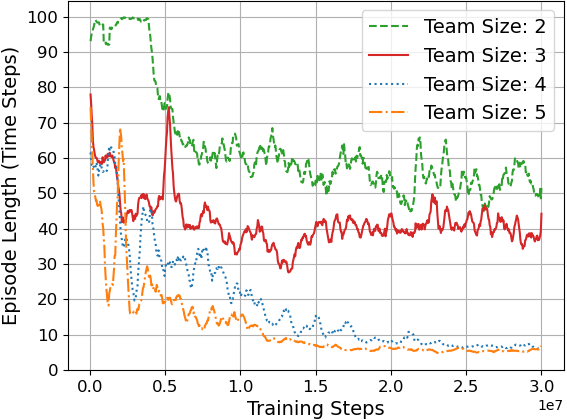}
         \caption{Target Strength: $1\times 10^8$}
         \label{ResultsTime3}
     \end{subfigure}

     \hfill
     \hfill

    \vspace{-0.5em}
    
    \begin{subfigure}{0.32\columnwidth}
         \centering
         \includegraphics[width=\linewidth]{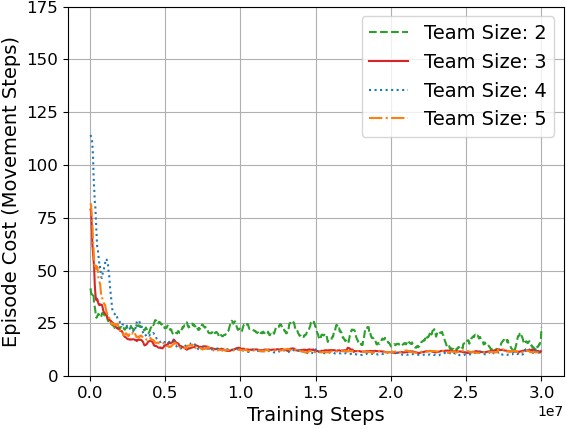}
         \caption{Target Strength: $1\times 10^9$}
         \label{ResultsCost1}
     \end{subfigure}
     \begin{subfigure}{0.32\columnwidth}
         \centering
         \includegraphics[width=\linewidth]{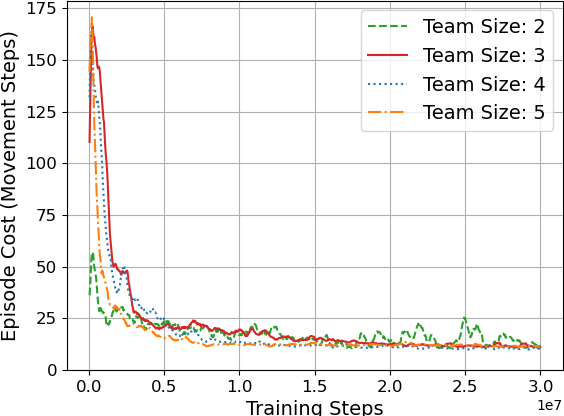}
         \caption{Target Strength: $5\times 10^8$}
         \label{ResultsCost2}
     \end{subfigure}
     \begin{subfigure}{0.32\columnwidth}
         \centering
         \includegraphics[width=\linewidth]{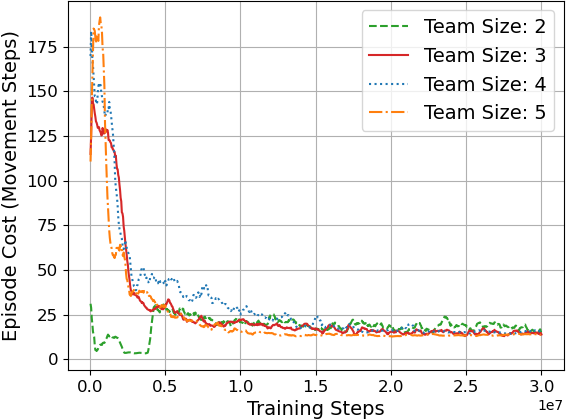}
         \caption{Target Strength: $1\times 10^8$}
         \label{ResultsCost3}
     \end{subfigure}

     \hfill
     \hfill
     \vspace{-1.5em}
     \caption{A summary of the training plots for different team sizes and for varying target strengths. The episodic reward is shown in (a)-(c), the episodic length is shown in (d)-(f), and the episodic cost is shown in (g)-(i).}
        \label{TrainingResults}
\end{figure}

Figure \ref{TrainingResults} summarizes the MADRL training results of the proposed method under different scenarios. Each sub-figure shows the results for varying team sizes to validate the scalability of the proposed approach. The figure also shows different target strengths, where $S$ varies from $1\times 10^9$ (strong) to $1\times 10^8$ (weak). A strong target can be detected from further away, and hence it is easier to detect and localize when compared to weaker target. It is worth mentioning that target estimation is not considered here yet since it is a different training process.

As seen in the figure, the training converges in all scenarios, indicating the feasibility of the proposed methods. When comparing the reward plot across different team sizes in Fig. \ref{ResultsReward1}, it can be seen that the 2-agent team performs the worst, while the other three team sizes perform better. This reflects the challenging nature of the problem for a team of only 2 agents, as it would require them more time to search the area. On the other hand, the similar performance across the other team sizes indicates that the agents efficiently learn to cooperate to finish the task in a timely manner with reduced resource consumption. The reason behind converging to negative rewards is because the agents take exploration steps in the early stages of an episode, which are necessary for data collection. In scenarios where a target exists, the agents initially do not have sufficient readings to take proper decisions and move towards the target, and hence could collaborate to move in the environment to collect data, which could initially give negative rewards if agents move away from the target. In cases where the target is behind obstacles, the agents might need to move more to collect data, resulting in more negative rewards. Nonetheless, once data are gathered, the agents immediately learn to make the right decisions, which explains the low magnitude of negative rewards (around -17) that the models converge to. This efficient performance, in terms of time and cost, can be seen in Fig. \ref{ResultsTime1} and Fig. \ref{ResultsCost1}. In terms of episode length, it can be noticed that bigger team sizes give lower search time, as expected. However, proving the efficacy of the proposed team-based reward function, it can be noticed that nearly all the team sizes (aside from team size = 2) converge to a similar episodic cost. This indicates that, while larger teams have higher expected cost per step, the efficient coordination and the timely search result in lower episodic cost. The agents make cost efficient decisions, such as staying idle if found non-beneficial to the team, resulting in lower resource consumption.

Studying the performance across different target strengths shows the increasing difficulty of the problem with lower target strengths. It should be noted that the range of target strengths used represents extreme scenarios varying from a target that is easily detectable from a long distance (i.e. strength: $1\times 10^9$) to one that is only detectable near the target (i.e. strength: $1\times 10^8$). It can be seen that, as the target strength increases, the difference in performance between the different team sizes increases, as smaller team sizes need to put more effort to finish the task. It is evident that all teams achieve a lower reward when going from higher target strength to a lower one, i.e. from Fig. \ref{ResultsReward1} to Fig. \ref{ResultsReward2} and from Fig. \ref{ResultsReward2} to Fig. \ref{ResultsReward3}. The two-agent and three-agent teams collect the lowest rewards for the low target strengths, indicating the need for bigger teams. This is evident when analyzing the episode length (Fig. \ref{ResultsTime1}-\ref{ResultsTime3}) and the episode cost (Fig. \ref{ResultsCost1}-\ref{ResultsCost3}). In terms of episodic length, more time is needed to search for weaker targets. This is mainly because weaker targets can be detected within a shorter range, hence requiring the agents to get closer. It can be also seen that the weakest target shows the need for larger teams for faster execution, since more search can be done in less time. When looking at the episodic cost, it is noticed that it increases when going from strong to weak targets, since more search is needed and hence more resource consumption. However, it is evident that regardless of the team size, the cost is nearly similar, with a slight advantage for bigger teams.

\begin{figure}[h]
     \centering
     \begin{subfigure}{0.48\columnwidth}
         \centering
         \includegraphics[width=\linewidth]{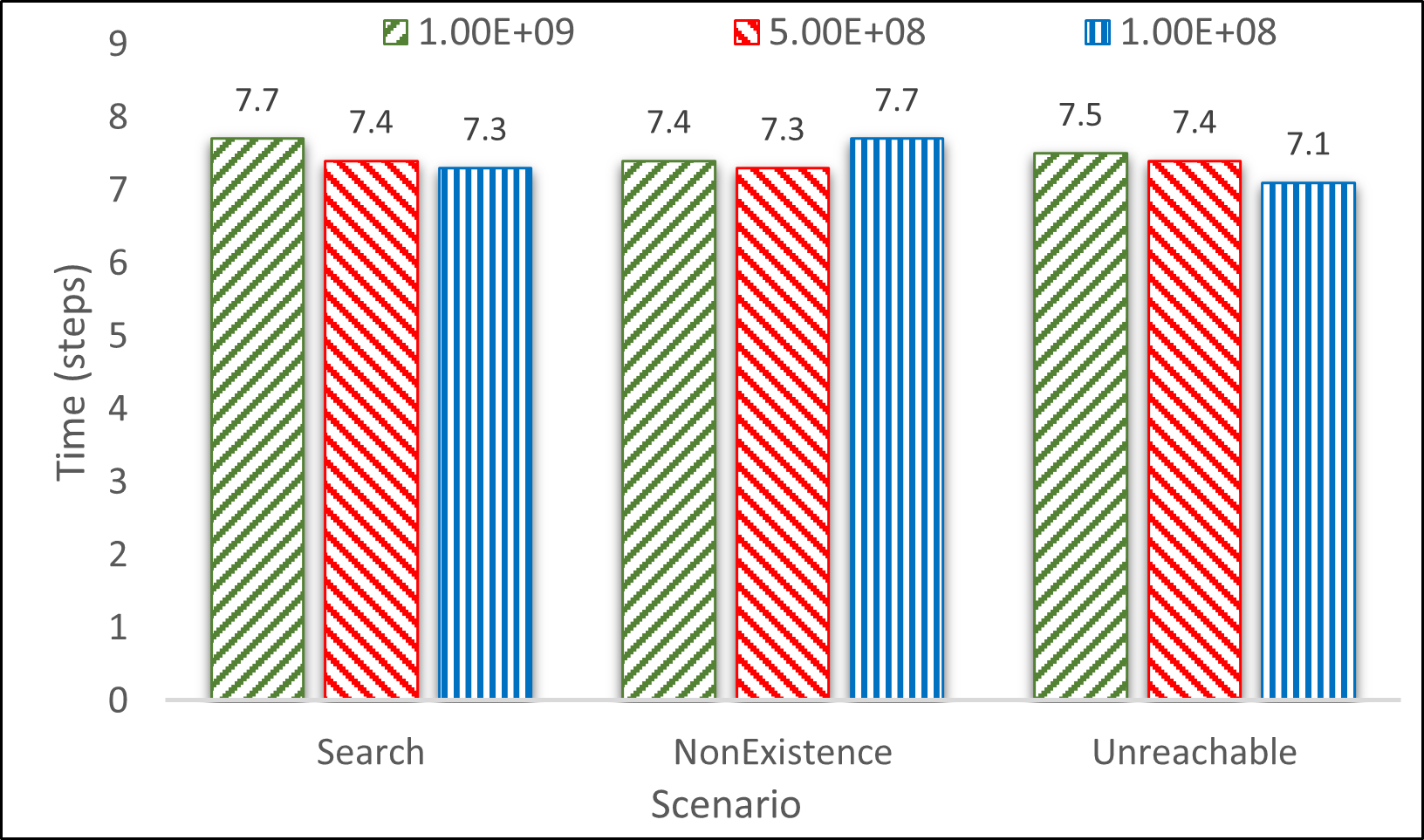}
         \caption{}
         \label{ScneariosResultsTime}
     \end{subfigure}
     \begin{subfigure}{0.48\columnwidth}
         \centering
         \includegraphics[width=\linewidth]{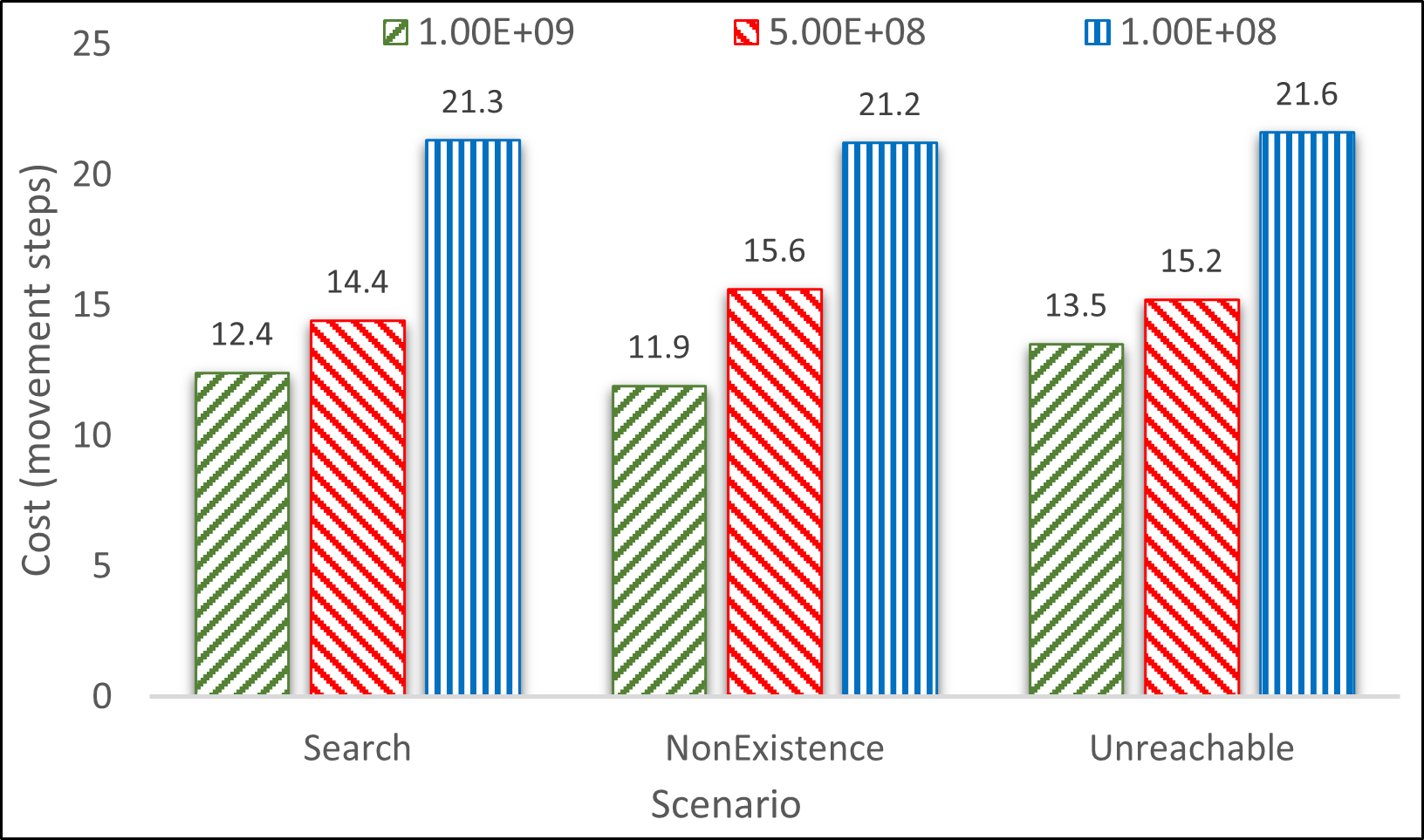}
         \caption{}
         \label{ScneariosResultsCost}
     \end{subfigure}
     
     \caption{The agents' performance under the different scenarios in terms of (a) episodic time and (b) episodic cost, for a team of 4 agents and varying target strengths.}
        \label{Results: Scenarios}
\end{figure}

To further study the model's behaviors, Fig. \ref{Results: Scenarios} analyzes the agents' performance under different scenarios, namely target search, target non-existence, and target unreachability, given a team of 4 agents under different target strengths. To obtain these results, the trained agents are deployed in the corresponding environment for 40,000 steps and the average performance is collected. As evident, in terms of time, the agents are equally able to reach the correct decision regardless of the scenario or the target strength. However, as evident in Fig. \ref{ScneariosResultsCost}, more cost is spent as the target strength gets weaker. This indicates that due to the increased difficulty of the process with weaker targets, more agents need to be involved to maintain a timely decision-making, resulting in higher costs. This behavior reflects the desired cooperation between the agents to achieve quick decision-making. Additionally, It can be noticed that the difference between time and cost is insignificant for stronger targets. For example, in the case of the strongest target, the search process on average takes 7.7 time steps and costs 12.4 movement steps in total. This means the agents quickly determine their informativeness to the task, and only informative agents carry the tasks while others maintain idle status to preserve resources.

\subsection{Target Estimation}
\begin{figure}[h]
    \centering
    \includegraphics[width=0.5\columnwidth]{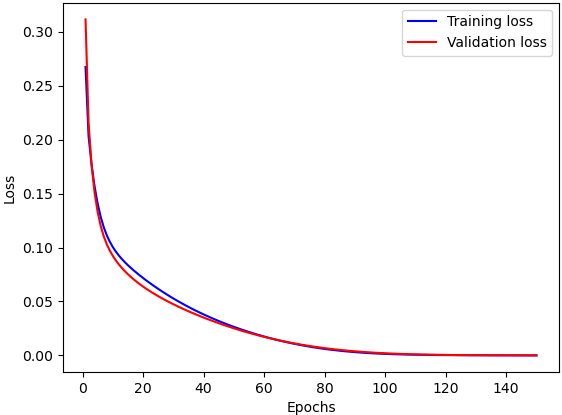}
    \caption{The training and validation loss for the target estimation model using TL.}
    \label{Results: Estimation}
\end{figure}

The aforementioned results analyze the performance of the MADRL model, which is responsible for the decision-making related to target search and flagging its non-existence or unreachability. Following the unreachability flag, a DL model is used to estimate the target locations based on the collected observations. Fig. \ref{Results: Estimation} shows the training loss and validation loss curves obtained while training a DL model, using TL, for target estimation. Here, a dataset is built combining the observations collected by the MADRL agents until the unreachability flag is produced, and then used to train the model to estimate the target. The dataset is split into training and validation sets, where the training set is used to update the model and the validation set is used to assess the model on unseen data. The training is done to reduce the Mean Squared Error (MSE) loss between the predicted (x,y) coordinates and the true coordinates. As can be seen in the figure, upon the completion of the training, both the training and validation loss values converge to 0, indicating that the model is able to accurately estimate the target location. One factor leading to these good results is the use of TL to transfer knowledge from the MADRL model to the DL model. Additionally, the observations collected by the MADRL agents and used to determine the unreachability of the target prove efficient in estimating its location.

\subsection{Benchmarks}
This section compares the performance of the proposed MADRL method with some existing benchmarks in the literature. As discussed in Section \ref{Sec: Intro}, existing methods struggle when addressing the realistic scenarios of target localization. To reflect on this, we compare the proposed approach with three existing benchmarks covering traditional target search techniques, single-agent DRL methods, and MADRL-based methods with no considerations to the uncertainties about the target. For the subsequent results, 4 agents are placed in a simulation environment with varying target strengths to analyze the episode length (time) and cost. After training, the agents are placed in different environments for inference, where the episode length and cost are averaged throughout 40k time steps. Agents obtained following these methods are placed in the environment described in Section \ref{subsection: Simulation Environment}, where the aim is to make decisions regarding the target location, its existence, and its reachability. Each episode has a limit of 100 steps, within which the agents are to finish the task. For fair comparison, the DRL- and MADRL-based solutions have been trained for the same number of steps as the proposed approach, i.e. for 30 million environment steps. The benchmarks are summarized as the following:

\begin{itemize}
    \item Uniform: A traditional method where agents follow a search path that is pre-defined to cover the area uniformly \cite{ziock2002lost}.
    \item DDQN: A single-agent DRL approach using Double Deep Q-learning for target search. The model is extrapolated into multi-agent settings by having a single model control all the agents in a centralized manner. The policy takes all the observations of the agents at once and produces a joint action vector \cite{liu2019double}.
    \item ODMTL: An optimized deep multi-agent target localization using MADRL \cite{alagha2022target, alagha2023multi}. Here, the methods follow a CLDE approach, but they do not consider scenarios of target unreachability or target non-existence.
\end{itemize}

\begin{figure}[h]
     \centering
     \begin{subfigure}{0.48\columnwidth}
         \centering
         \includegraphics[width=\linewidth]{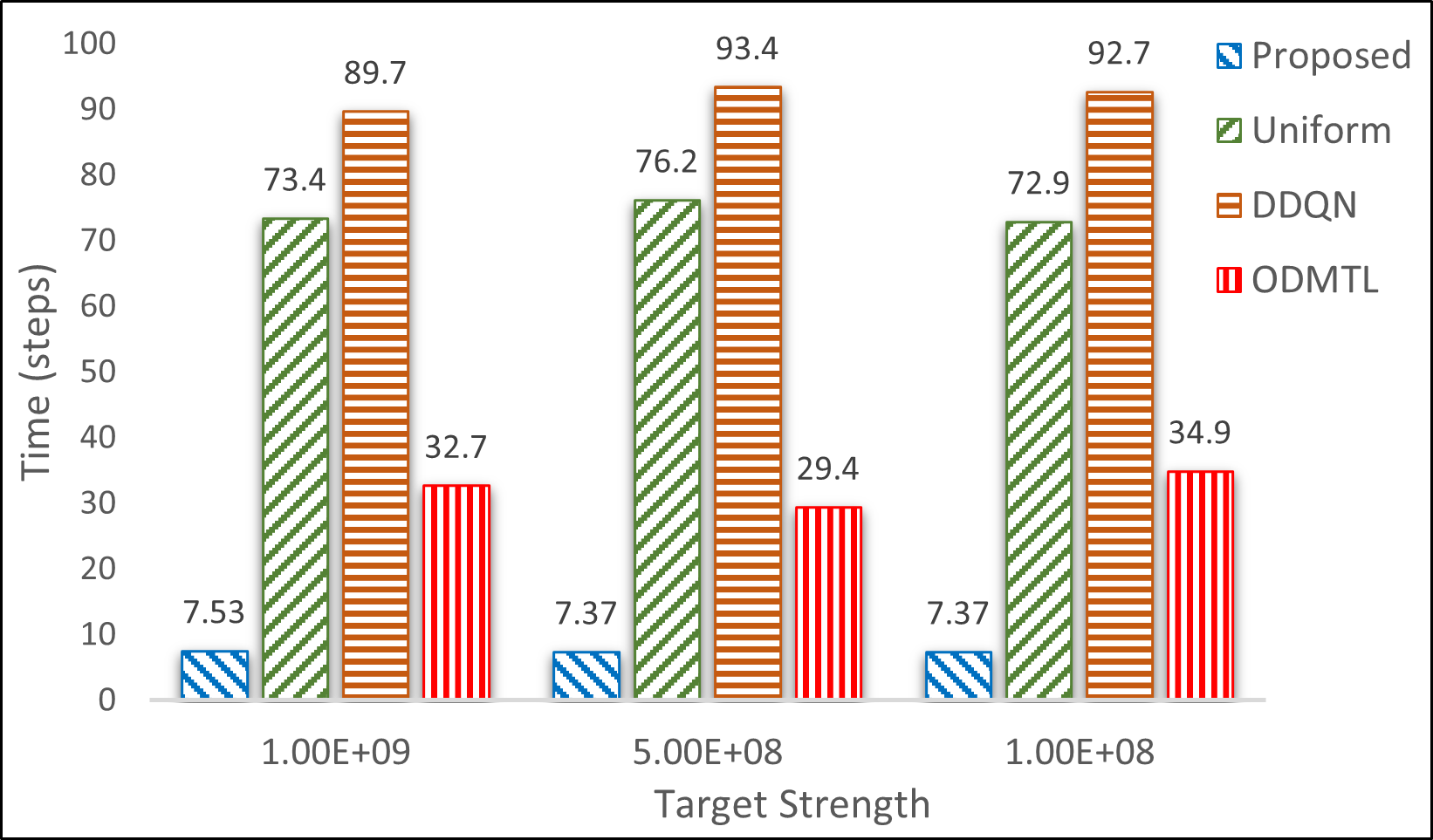}
         \caption{}
         \label{BenchmarksTime}
     \end{subfigure}
     \begin{subfigure}{0.48\columnwidth}
         \centering
         \includegraphics[width=\linewidth]{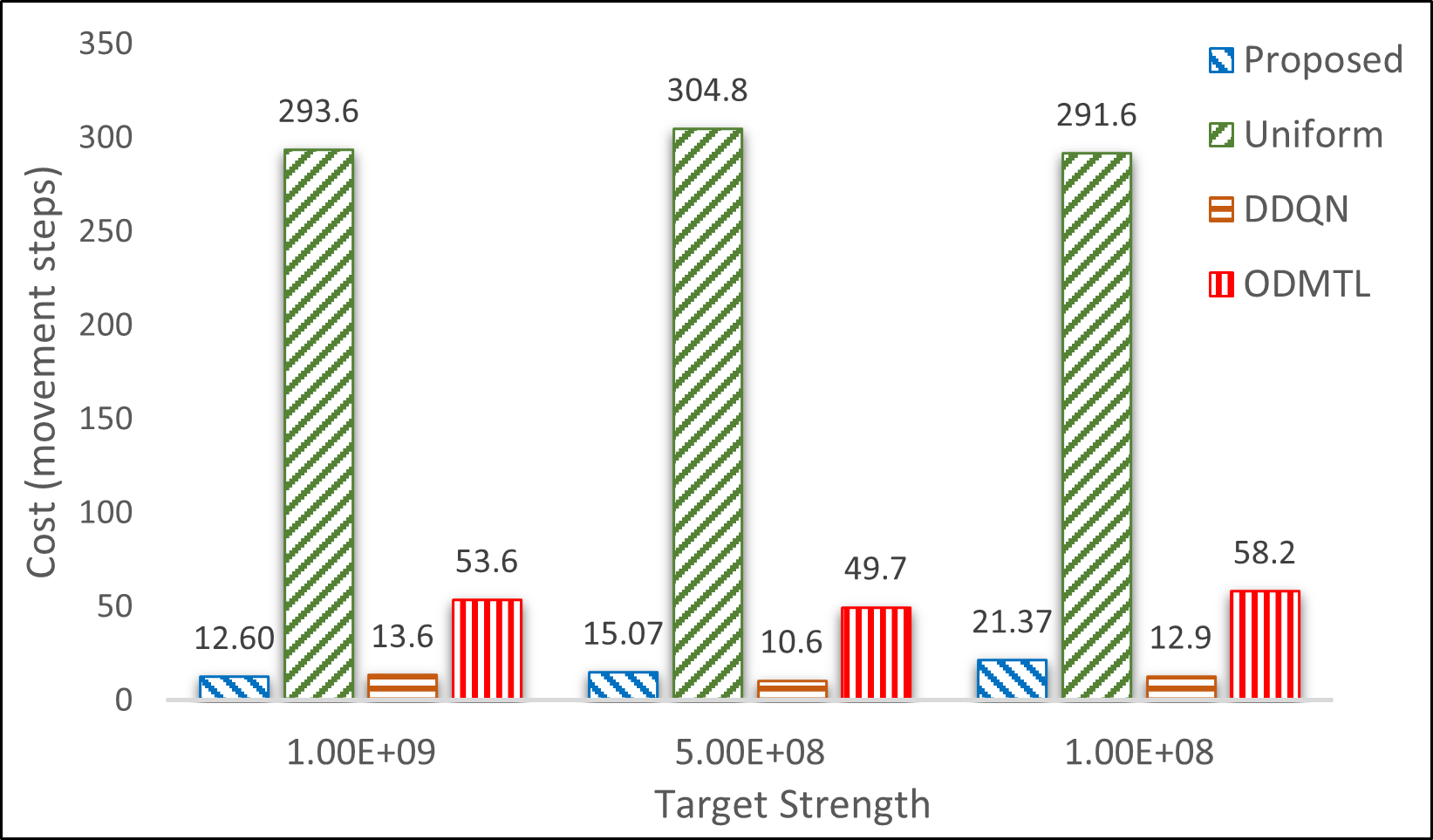}
         \caption{}
         \label{BenchmarksCost}
     \end{subfigure}
     
     \caption{Comparison between the performance of the proposed method and the benchmarks in terms of (a) episode length and (b) episode cost.}
        \label{Results: Benchmark}
\end{figure}

Figure \ref{Results: Benchmark} summarizes the performances of the different benchmarks under varying target strengths in terms of time and cost. As can be seen, the proposed work outperforms all the benchmarks by achieving faster and less costly localization tasks. Traditional uniform search methods do guarantee finding the target, but cannot handle cases where the target is unreachable or does not exist. In such scenarios, the agents keep searching the environment and waste resources. Additionally, due to the lack of cooperation between agents, there is an increased resource consumption. When comparing the proposed work with single-agent DDQN methods, it can be seen that these methods struggle to learn to perform the task. Here, within the same amount of learning experience, the agents fail to learn any desired behaviors and resort to staying idle most of the time, hence the high localization time and low cost. This is a form of local optima, which is a result of the scalability issues faced in single-agent DRL methods when extrapolated into multi-agent settings. The ODMTL method generally performs better than the other two methods, due to the existence of intelligent decision-making and cooperation between agents. However, the lack of consideration of the uncertainties in the environment, such as the non-existence of the target or its unreachability, hinder their ability.

\subsection{Complexity Analysis}
This section aims to analyze the time and space complexity of the proposed methods while also comparing them with the benchmarks. Here, we compare the proposed approach with the previously discussed two DRL-based methods, i.e. DDQN \cite{liu2019double} and ODMTL \cite{alagha2023multi}. Table \ref{StateActionDims} shows the dimensionality of the state and action spaces for the different methods as a function of the number of agents $N$ and the input size $I$. The input size here reflects the number of pixels per observation. For example, in our method, each of the 9 "reduced" observations has a $7\times 7$ dimensionality, resulting in $I=441$. While the number of observations is fixed (9), the dimensionality of the observations is the only variable affecting the complexity. As can be seen in the table, for ODMTL and the proposed method, the dimensionality of the state/action spaces is not a function of $N$, showing their  scalability as the number of agents increases. This is mainly due to the distributed nature of these methods, where each agent acts independently in its own action/state space. The main difference between the two methods lies in the action space, where ODMTL has no considerations for target uncertainties, resulting in a slightly less complex actions but worse performance, as shown previously in Fig. \ref{Results: Benchmark}. When considering both state and action complexities, the total space complexity of the two algorithms is a linear function of $I$ (i.e. $O(I)$), which constant for a given problem and does not depend on the number of agents. On the other hand, the space and actions spaces of the centralized DDQN method rely heavily on the number of agents. For each new agent, the centralized actor adds two additional reduced observations for their location. More importantly, the action space exponentially increases with each agent, as the centralized actor operates in the combined action space that covers all the possible combinations of actions between the agents. This results in an exponential space complexity ($O(11^N)$) for this algorithm. The complexity of the state/action spaces can be reflected in the number of trainable parameters for each method, as seen in the table. For the distributed methods, i.e. ODMTL and the proposed method, the number of trainable parameters is fixed regardless of the team size. On the other hand, for the centralized DDQN method, the number of trainable parameters explodes with more agents, indicating its scalability issues. In summary, the proposed method maintains comparable scalability, if not better, when compared to existing benchmarks, with the added advantage of outperforming them for the problem of target localization in uncertain environments.

\begin{table}[ht]
\caption{The number of trainable parameters as well as the space complexity for the different methods.}
\vspace{-1em}
\setlength{\tabcolsep}{0pt}
\begin{center}
\begin{tabular}{|P{0.12\columnwidth}|P{0.085\columnwidth}|P{0.085\columnwidth}|P{0.085\columnwidth}|P{0.085\columnwidth}|P{0.14\columnwidth}|P{0.16\columnwidth}|P{0.23\columnwidth}|}
\hline
\rowcolor{Gray} & \multicolumn{4}{c|}{$\sim$Trainable Parameters per $N$} &  & & \\
\cline{2-5}
\rowcolor{Gray} \multirow{-2}{*}{Model} & 2 & 3 & 4 & 5 & \multirow{-2}{*}{State Space} & \multirow{-2}{*}{Action Space} & \multirow{-2}{*}{Space Complexity}\\
\hline
DDQN & \multicolumn{1}{c|}{359k} & \multicolumn{1}{c|}{442k} & \multicolumn{1}{c|}{1195k} & \multicolumn{1}{c|}{7966k}& \multicolumn{1}{c|}{$(2N+7)I$}& \multicolumn{1}{c|}{$11^N$}& \multicolumn{1}{c|}{$O(11^N)$}\\
\hline
ODMTL & \multicolumn{4}{c|}{350k}& \multicolumn{1}{c|}{$9I$}& \multicolumn{1}{c|}{$9$}& \multicolumn{1}{c|}{$O(I)$}\\
\hline
Proposed & \multicolumn{4}{c|}{350k}& \multicolumn{1}{c|}{$9I$}& \multicolumn{1}{c|}{$11$}& \multicolumn{1}{c|}{$O(I)$}\\

\hline

\hline
\end{tabular}

\end{center}
\label{StateActionDims}
\vspace{-1em}
\end{table}

In terms of time complexity, RL algorithms relying on Temporal Difference (TD) error have a time complexity of $O(d)$, where $d$ is the complexity of the model used (i.e. the neural network) \cite{sutton2018reinforcement}. PPO and DQN use TD in batches and epochs, resulting in a complexity of $O(K \times B\times d)$, where $K$ is the number of epochs and $B$ is the batch size used. Since $K$ and $B$ are hyperparameters, the main influencer on the time complexity here would be the complexity of the policy model $d$. To assess this complexity, we use the FLOPs (Floating Point Operations) metric, which is commonly used in practice to estimate the computational cost of training DL and  DRL models, and which can be easily translated into training speed and power consumption depending on the machine used. The FLOPs metric is computed by tracking the total number of floating-point operations by all the layers of a DL model. In this context, both ODMTL and the proposed method have approximately 1 MFLOPs ($10^6$) based on the CNN architectures used. This is regardless of the number of agents, which shows their scalability. On the other hand, the FLOPs for the centralized DDQN method range from 1 MFLOPs for 2 agents all the way to 41 MFLOPs for 5 agents, which shows the method's scalability issues. Generally, the proposed method uses a simple policy architecture that is efficient for real-time applications. For reference, commonly used architectures like ResNet-18 and MobileNet-V2 have 1.82 GFLOPs and 319 MFLOPs \footnote{\href{https://mmclassification.readthedocs.io/en/latest/model_zoo.html}{https://mmclassification.readthedocs.io/en/latest/model\_zoo.html}}, respectively, which are much more complex than our proposed method. In terms of inference, the proposed model takes on our machine less than 1 ms to produce an action based on the observations of an agent. 

While the training time for the proposed method depends significantly on the used machine, a trade-off analysis can be conducted to assess the effect of the number of agents and the target strength on the training process. Specifically, training the 30M environment steps shown in Fig. \ref{TrainingResults} takes 22.3h on average regardless of the team size. This is due to the fact that there is no increase in time complexity in the decision-making process (during training and inference), since agents act independently in parallel. PPO allows for parallelized experience collection, which makes the process scalable. On the other hand, some variance in results can be noticed in terms of the convergence time, based on the number of agents and target strength. For the same target strength, earlier convergence can be noticed with higher group sizes, as seen in Fig. \ref{TrainingResults}. Furthermore, when studying one group size across different target strengths, it can be seen that the models take longer to converge as the target gets weaker. This is mainly because weaker targets need more search effort to be found since they are harder to detect. It is worth mentioning that the aforementioned training time could be further reduced with access to more resources through parallelized processing. For reference, the team from OpenAI in \cite{baker2019emergent} needed up to 99h to train their MADRL system, despite the use of powerful computational resources. Generally, since training is done offline, it is feasible for the training process to take such long times, as long as the inference of the model after training is time-efficient, which is the case of our model as shown previously.

\section{Conclusion}
\label{Conclusion}
In this paper, the target localization problem is addressed using MADRL methods, while considering realistic scenarios of target non-existence and unreachability. Based on the collected observations, which are modeled as 2D maps representing the environment, the agents learn efficient decision-making through MADRL, encapsulating actions related to the mobility in the environment and determining the existence and reachability of the target. The MADRL policy is represented by a CNN which is optimized using PPO and a team-based shaped reward function. Using TL, the same model is expanded to also cover target estimation, where the target coordinates are approximated if it is unreachable. The proposed MADRL method was tested on different scenarios, covering varying target strengths (from weak to strong) and varying number of agents. The scalability and adaptability of the proposed method was verified through quick and efficient target localization tasks, with very accurate target estimation. Compared to traditional and DRL-based benchmarks, the proposed work shows outperformance, especially in tackling scenarios with uncertainties such as target non-existence and unreachability.

Future directions could extend the current approach to handle multi-target localization, which presents additional challenges and requires modifications and new formulation of the observations, policy structure, and reward mechanisms to coordinate the agents in locating multiple targets. Additionally, a promising direction would be to expand the method to handle mobile targets, adding another layer of complexity to the agents’ decision-making processes. This would involve developing dynamic observations and adaptive strategies to enable agents to track and localize moving targets effectively in dynamic environments.

\bibliographystyle{model1-num-names}
\bibliography{bibliography}
\end{document}